\documentclass[10pt,twocolumn,letterpaper]{article}

\usepackage{iccv}
\usepackage{times}
\usepackage{epsfig}
\usepackage{graphicx}
\usepackage{amsmath}
\usepackage{amssymb}
\usepackage{booktabs}
\usepackage{amstext}
\usepackage{url}            
\usepackage{amsfonts}       
\usepackage{nicefrac}       
\usepackage{microtype}      
\usepackage{multirow}
\usepackage[table,xcdraw]{xcolor}
\usepackage{bm}
\usepackage{amsmath}
\usepackage{amssymb}
\usepackage{paralist}
\usepackage{wrapfig}
\usepackage{bbding}
\usepackage{multicol}
\usepackage{color}
\usepackage{pifont}
\usepackage{multirow}
\usepackage{booktabs}
\usepackage[accsupp]{axessibility} 

\usepackage[pagebackref=true,breaklinks=true,letterpaper=true,colorlinks,bookmarks=false]{hyperref}

\iccvfinalcopy 

\def\iccvPaperID{2670} 

\ificcvfinal\fi

\begin{document}

\title{Multi-grained Temporal Prototype Learning for Few-shot Video Object Segmentation}

\author{
    Nian Liu$^{1}$\enspace 
    Kepan Nan$^{2}$\enspace 
    Wangbo Zhao$^{3}$\enspace 
    Yuanwei Liu$^{2}$\enspace 
    Xiwen Yao$^{2}$\footnotemark[2] \\
    Salman Khan$^{1,4}$\enspace
    Hisham Cholakkal$^{1}$\enspace
    Rao Muhammad Anwer$^{1}$\enspace
    Junwei Han$^{2}$\enspace 
    Fahad Shahbaz Khan$^{1,5}$ 
    \\
    $^1$ Mohamed bin Zayed University of Artificial Intelligence \enspace 
    $^2$ Northwestern Polytechnical University \enspace 
    \\ 
    $^3$ National University of Singapore
    \\
    $^4$Australian National University\enspace 
    $^5$CVL, Linköping University
}

\maketitle
\footnotetext[2]{Corresponding author: yaoxiwen517@gmail.com.}
\ificcvfinal\thispagestyle{empty}\fi

\begin{abstract}
   Few-Shot Video Object Segmentation (FSVOS) aims to segment objects in a query video with the same category defined by a few annotated support images. However, this task was seldom explored. In this work, based on IPMT, a state-of-the-art few-shot image segmentation method that combines external support guidance information with adaptive query guidance cues, we propose to leverage multi-grained temporal guidance information for handling the temporal correlation nature of video data. We decompose the query video information into a clip prototype and a memory prototype for capturing local and long-term internal temporal guidance, respectively. Frame prototypes are further used for each frame independently to handle fine-grained adaptive guidance and enable bidirectional clip-frame prototype communication. 
   To reduce the influence of noisy memory, we propose to leverage the structural similarity relation among different predicted regions and the support for selecting reliable memory frames. Furthermore, a new segmentation loss is also proposed to enhance the category discriminability of the learned prototypes. Experimental results demonstrate that our proposed video IPMT model significantly outperforms previous models on two benchmark datasets. Code is available at \href{https://github.com/nankepan/VIPMT}{https://github.com/nankepan/VIPMT}.
\end{abstract}

\section{Introduction}
To mitigate the data-hungry issue of modern deep semantic segmentation models \cite{long2015fcn,chen2017deeplab,cheng2021maskformer}, few-shot semantic segmentation emerges by only requiring a few support samples with annotated masks for segmenting the objects of the same class in new images. Many recent works \cite{tian2020pfenet,zhang2021cyctr,lang2022bam,liu2022ipmt} have shown very promising results on image data using the meta-learning scheme. They simulate the inference process and partition the training set into numerous episodes, in each of which, the model samples a few support images and learns to guide the segmentation on the query images.

\begin{figure}[t]
\centering
    \includegraphics[width=1\linewidth]{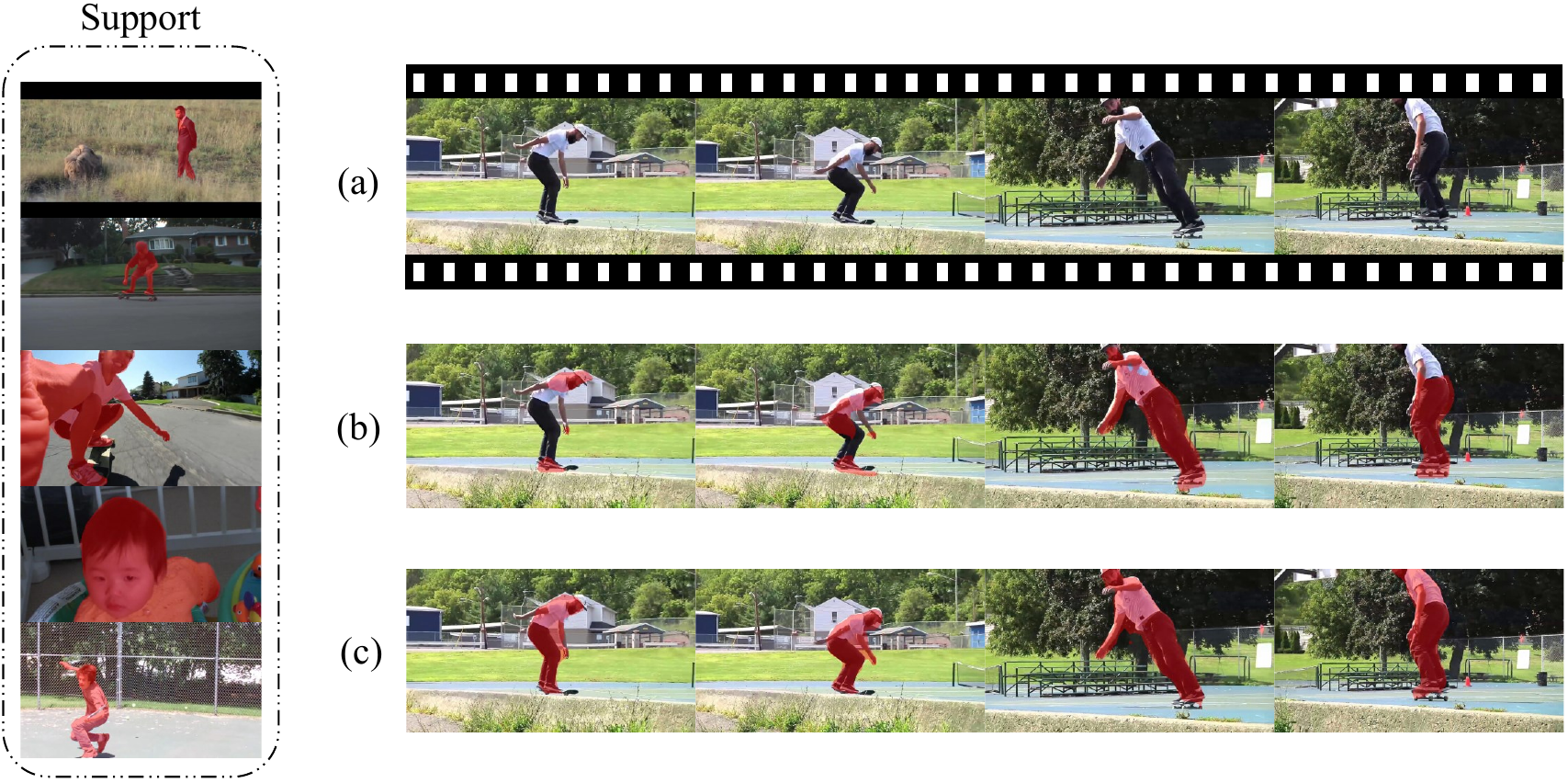}
	\caption{Given a few annotated support frames, FSVOS aims to segment the objects with the same category from a query video (a). Simply considering single-frame information leads to inconsecutive segmentation results (b), while our method generates accurate results by using multi-grained temporal prototypes (c).}
	\label{fig:intro}
	\vspace{-3mm}
\end{figure}

Inspired by the classic Few-Shot Image Semantic Segmentation (FSISS) task, the Few-Shot Video Object Segmentation (FSVOS) task was introduced by \cite{chen2021danet,siam2022tti}.
For FSISS, many works adopt the prototype-based methods, in which a prototype vector is extracted from the support to encode the category guidance information, and then a segmentation head learns to match the prototype with the feature at each query pixel for performing query segmentation. However, the intra-class diversity can cause the matching gap between the support-induced prototype guidance and the query features.
The IPMT model \cite{liu2022ipmt} solved this problem by learning an intermediate prototype that integrates both \emph{support-induced external category guidance knowledge} and \emph{query-induced adaptive guidance information}.
As for FSVOS, video data additionally show temporal correlation, from which we can also induce \emph{internal temporal guidance}.
If we ignore this prior knowledge and simply consider single-frame information, the learned prototypes may vary significantly among different frames, leading to inconsecutive segmentation results, as shown in Figure~\ref{fig:intro} (b).

In this work, we extend the IPMT model \cite{liu2022ipmt} for video data by tackling the temporal correlation nature. Based on the intermediate prototype mechanism of IPMT, we propose to decompose the query video information into the clip-level, frame-level, and memory-level prototypes, consisting of a multi-grained temporal structure, which was \emph{NEVER} explored by existing models \cite{chen2021danet,siam2022tti}. Among them, the clip prototype encodes local temporal object guidance information in each consecutive clip, while the memory prototype introduces long-term historical guidance cues. Combing them can effectively handle the temporal correlation problem. 
However, such a design may ignore fine-grained per-frame adaptive guidance information, hence may fail to handle large scene changes and object transformation.
To this end, we further learn an adaptive frame prototype by independently encoding per-frame object features. Additionally, we enable bidirectional clip-frame prototype communication by making the clip-level and frame-level prototypes initialize each other, hence promoting intra-clip temporal correlation.

To better leverage the historical memory guidance, we also follow a Video Object Segmentation (VOS) method \cite{liu2022learning} and train an IoU regression network to select reliable memory frames for reducing the negative influence brought by noisy memory. However, different from \cite{liu2022learning}, we explicitly consider the nature of the FSVOS task and propose to leverage the structural similarity relation among different predicted regions and the support information for predicting more accurate IoU scores. To make the learned prototype more category discriminative, we also propose a Cross-Category Discriminative Segmentation (CCDS) loss by leveraging negative batch samples. Extensive experimental results have verified the effectiveness of our proposed \textbf{Video IPMT} (VIPMT) model and showed its significant performance improvement over state-of-the-art models.

In conclusion, our contributions can be summarized as follows:
\begin{compactitem}
\item For the very first time, we propose to learn multi-grained temporal prototypes for FSVOS, by extending the IPMT \cite{liu2022ipmt} model. Clip and memory prototypes are learned for internal temporal guidance. Frame prototypes are used for fine-grained adaptive guidance and also enable prototype communication.
\item We propose to leverage the structural similarity relation among different predicted regions and the support for selecting reliable memory information. We also present a CCDS loss using the negative samples within each batch for promoting category discriminability of the learned prototypes.
\item Experimental results have demonstrated the significant effectiveness of our proposed model, which improves state-of-the-art results by more than 4\% and 3\%, on two benchmark datasets, respectively.

\end{compactitem}

\section{Related Work}
\subsection{Few-Shot Image Semantic Segmentation}
Most previous FSISS works adopted meta-learning-based methods \cite{vinyals2016matching}, especially 
prototype-based models. Specifically, Dong and Xing \cite{dong2018few} aggregated a prototype vector on the support images to encode the representative category knowledge first, and then evaluated its similarity with each query pixel in a matching network as the segmentation result. Following this idea, 
Yang \etal \cite{yang2020pmm} constructed multiple prototypes from limited support images to represent diverse image regions. 
In \cite{tian2020pfenet}, Tian \etal used high-level features to generate a prior mask as a supplement to the support prototype for refining the query feature.
Liu \etal \cite{liu2022ntrenet} leveraged non-target prototypes to eliminate distracting regions. BAM \cite{lang2022bam} used the prototypes of base classes to explicitly suppress corresponding regions. In IPMT \cite{liu2022ipmt}, Liu \etal proposed to generate an intermediate prototype
which encodes the category guidance information from both support and query. 
However, all these methods only focused on image data and the challenge of video data remains seldom explored.

\subsection{Video Object Segmentation}
Another closely related domain is VOS, especially the semi-supervised VOS, in which the mask label of an object in the first frame is given and the model is required to segment the same object in other frames. 
Some methods \cite{perazzi2017learning, cheng2017segflow, xiao2018monet} tried to leverage optical flow to help segment target objects.
Noticing the successful application of memory networks \cite{chunseong2017attend, yang2018learning} in computer vision, STM \cite{oh2019video} proposed a memory mechanism to leverage information from previous frames for segmenting the current frame. Lu \etal \cite{lu2020video} followed this idea and improve the memory mechanism with a graph model. 
For a comprehensive survey please refer to \cite{zhou2022survey}.

Different from semi-supervised VOS, we tackle the FSVOS task with two differences. First, semi-supervised VOS aims to segment \emph{the same object} indicated in the annotation of \emph{the first frame} while FSVOS requires to segment \emph{the objects of the same class} with the annotated support set.
Second, the support set in FSVOS 
can be composed of \emph{any images or frames of any videos} that contain the target class. Hence, FSVOS is more generalized and faces much larger intra-class diversity between the support set and the query video. In this work, we propose multi-grained temporal prototype learning and bidirectional clip-frame prototype communication for FSVOS, which is different from existing VOS methods.

Our idea of using the IoU network to select memory frames with high segmentation quality is inspired by \cite{liu2022learning}. However, in \cite{liu2022learning}, the authors directly regress the IoU score by only taking the image and the mask as the network input.
In this work, 
we further consider the nature of FSVOS and propose to compute several structural similarity maps which explicitly encode the quality assessment prior of the relations among the predicted foreground, background, and the support areas.

\subsection{Few-shot Video Object Segmentation}
For the FSVOS task, currently, only a few works have targeted this topic. Chen \etal \cite{chen2021danet} proposed the first FSVOS dataset and model. They proposed a Domain Agent Network (DAN) to alleviate the large computational cost of the many-to-many attention between the support images and the query video frames, which only considered clip-level temporal information. Siam \etal \cite{siam2022tti} proposed a temporal transductive inference model, which uses both global and local temporal constraints to obtain per-frame model weights and locally consistent predictions.
Compared with them, we optimize multi-grained temporal prototypes while they optimized per-frame model weights. Furthermore, we explicitly use historical memory while they did not.
\begin{figure*}[htbp]
	\begin{center}
		\includegraphics[width=1\linewidth]{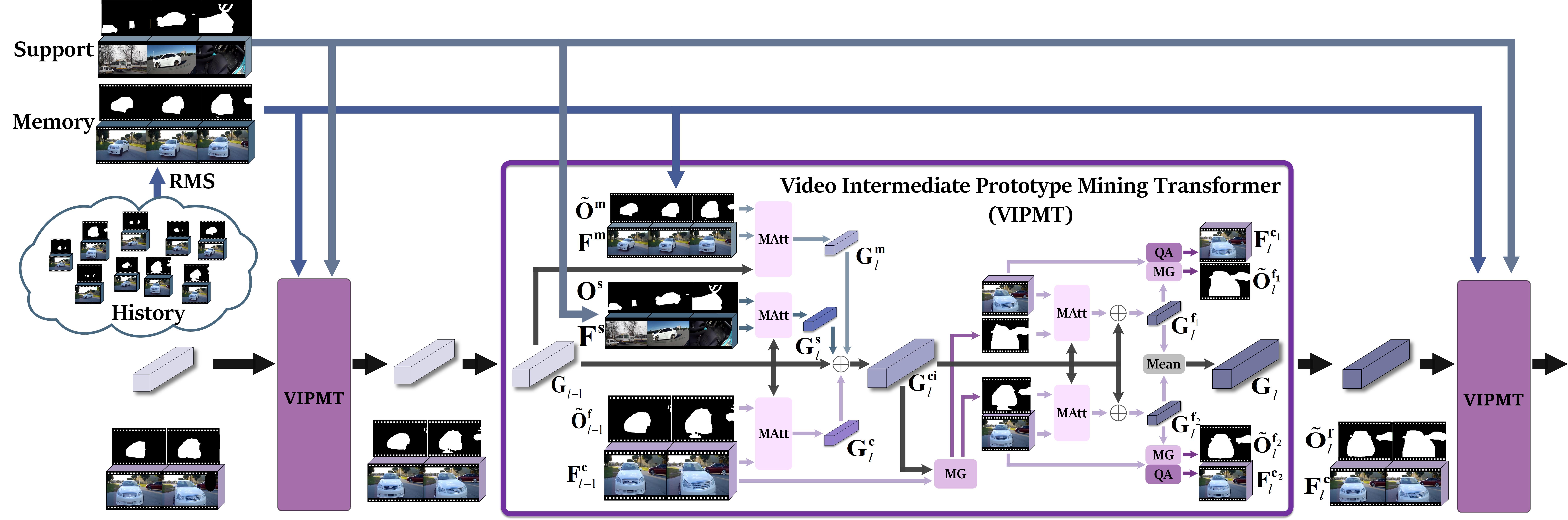}
	\end{center}
	\caption{\textbf{Main architecture of the proposed VIPMT model for FSVOS.} ``RMS'' means our proposed reliable memory selection (Section~\ref{memory prototype learning}), ``MAtt'' means masked attention \eqref{matt}, ``MG'' means mask generation \eqref{pred_func}, and ``QA'' means query activation \eqref{query_act}. In each VIPMT iteration, the input intermediate prototype $\mathbf{G}_{l-1}$ first generates the support prototype $\mathbf{G}^{\mathbf{s}}_{l}$ \eqref{g_support_query}, the clip prototype $\mathbf{G}^{\mathbf{c}}_{l}$ \eqref{g_clip}, and the memory prototype $\mathbf{G}^{\mathbf{m}}_{l}$ \eqref{g_mem}, and combine them to obtain the clip-level intermediate prototype $\mathbf{G}_{l}^{\mathbf{ci}}$ \eqref{ipm_clip_mem}. Afterwords, a frame level prototype $\mathbf{G}_{l}^{\mathbf{f}_i}$ is obtained for each frame via \eqref{g_frame} and then updates the frame mask $\tilde{\mathbf{O}}^{\mathbf{f}_i}_{l}$ and frame feature  $\mathbf{F}^{\mathbf{c}_i}_{l}$ via \eqref{frame_pred_qa}. Finally, all $\mathbf{G}_{l}^{\mathbf{f}_i}$ are averaged to obtain $\mathbf{G}_{l}$ \eqref{g_frame2clip}, which is input into the next iteration with $\tilde{\mathbf{O}}^{\mathbf{f}}_{l}$ and $\mathbf{F}^{\mathbf{c}}_{l}$.}
	\label{fig:overview}
	\vspace{-3mm}
\end{figure*}

\section{Preliminaries}

\subsection{Problem Definition}
For FSVOS, a whole video object segmentation dataset $\mathcal D$ with multiple object categories $\mathcal C$ is divided into a training subset $\mathcal D_{base}$ and a testing subset $\mathcal D_{novel}$, whose categories are $\mathcal C_{base}$  and $\mathcal C_{novel}$, respectively, and satisfy $\mathcal C_{base}\cap \mathcal C_{novel}=\varnothing$ and $\mathcal C_{base}\cup \mathcal C_{novel}=\mathcal C$. Under the standard few-shot setting and the episodic training scheme, both $\mathcal D_{base}$ and $\mathcal D_{novel}$ are randomly partitioned into a lot of episodes. In each episode, 
a support set $\mathcal S$ provides $K$ frames $\mathbf{I}^{\mathbf{s}}$ with labeled segmentation masks $\mathbf{O}^{\mathbf{s}}$ of a specific object category (we follow \cite{chen2021danet,siam2022tti} to focus on one-way segmentation here), \ie, $\mathcal S=\{(\mathbf{I}^{\mathbf{s}_i},\mathbf{O}^{\mathbf{s}_i})\}_{i=1}^K$, as the meta knowledge of the target class. Then, given a query video $\mathcal Q=\{\mathbf{I}^{\mathbf{q}_i}\}_{i=1}^N$ with $N$ frames, 
an FSVOS model is used to predict the segmentation masks $\{{\tilde{\mathbf{O}}}^{\mathbf{q}_i}\}_{i=1}^N$ of the same category for each frame.
The model can be trained on the episodes sampled from $\mathcal D_{base}$ with the given ground truth query masks $\{{\mathbf {O}}^{\mathbf{q}_i}\}_{i=1}^N$, learning how to transfer the guidance knowledge from the support to the query. Finally, the FSVOS model is expected to make accurate predictions for unseen categories on $\mathcal D_{novel}$.

\subsection{Review of IPMT}
The IPMT model \cite{liu2022ipmt} adopts a transformer-based architecture for the FSISS task. 
Specifically, FSISS can be modeled by learning an intermediate prototype $\mathbf{G} \in \mathbb{R}^{1\times C}$ and pixel-wise feature maps $\mathbf{F}^{\mathbf{s}}$ and $\mathbf{F}^{\mathbf{q}}$ for the support and query, respectively, and then conduct iterative mutual optimization between $\mathbf{G}$ and $\mathbf{F}^{\mathbf{q}}$ within multiple IPMT layers. 
Under the $K$-shot setting, the support features $\mathbf{F}^{\mathbf{s}} \in \mathbb{R}^{Khw\times C}$ and the flattened masks $\mathbf{O}^{\mathbf{s}} \in \mathbb{R}^{Khw\times 1}$, where $h,w,C$ are the height, width, and channel number, respectively. 
In the $l$-th IPMT layer, given the previous query feature $\mathbf{F}^{\mathbf{q}}_{l-1} \in \mathbb{R}^{hw\times C}$ and the flattened binarized mask prediction $\tilde{\mathbf{O}}^{\mathbf{q}}_{l-1} \in \mathbb{R}^{hw\times 1}$,
the intermediate prototype $\mathbf{G}$ is updated by
\begin{equation}
\label{ipm}
\begin{aligned}
        \mathbf{G}_{l}
        &= \mathbf{IPM}(\mathbf{G}_{l-1},\mathbf{F}^{\mathbf{s}},\mathbf{F}^{\mathbf{q}}_{l-1},\mathbf{O}^{\mathbf{s}},\tilde{\mathbf{O}}^{\mathbf{q}}_{l-1}) \\
        &=  \mathbf{MLP}(\mathbf{G}^{\mathbf{s}}_{l} + \mathbf{G}^{\mathbf{q}}_{l}+\mathbf{G}_{l-1}),
\end{aligned}
\end{equation}
where $\mathbf{MLP}$ denotes a multi-layer perception. The $\mathbf{G}^{\mathbf{s}}_{l}$ can be seen as the \emph{support prototype} which encodes the deterministic guidance knowledge from the support and the $\mathbf{G}^{\mathbf{q}}_{l}$ can be regarded as the \emph{query prototype} that learns adaptive guidance information from the query. They can be obtained by
\begin{eqnarray}
\label{g_support_query}
        \mathbf{G}^{\mathbf{s}}_{l} & = &  \mathbf{MAtt}(\mathbf{G}_{l-1},\mathbf{F}^{\mathbf{s}},\mathbf{O}^{\mathbf{s}}),\\
        \mathbf{G}^{\mathbf{q}}_{l} & = &  \mathbf{MAtt}(\mathbf{G}_{l-1},\mathbf{F}^{\mathbf{q}}_{l-1},\tilde{\mathbf{O}}^{\mathbf{q}}_{l-1}).
\end{eqnarray}
Here, $\mathbf{MAtt}$ means the masked attention operation \cite{cheng2022mask2former}:
\begin{equation}
\label{matt}
    \mathbf{MAtt}(\mathbf{G},\mathbf{F},\mathbf{O}) = \delta(f_Q(\mathbf{G})f_K(\mathbf{F})^{\top}+\Delta)f_V(\mathbf{F}),
\end{equation}
where $\delta$ means the softmax normalization, $f_Q(\cdot),f_K(\cdot),f_V(\cdot)$ are three linear transformations following \cite{vaswani2017attention}, and $\Delta=(1-\mathbf{O}^{\top})\cdot(-\infty)$ is used to modulate the attention matrix, 
making background attention weights become zeros after softmax.

Then, the updated $\mathbf{G}_{l}$ is used to generate mask predictions for both support and query via a mask generation ($\mathbf{MG}$) process:
\begin{equation}
\label{pred}
        \tilde{\mathbf{O}}^{\mathbf{s}}_{l} =  \mathbf{MG}(\mathbf{G}_{l}, \mathbf{F^s}),
        \tilde{\mathbf{O}}^{\mathbf{q}}_{l} =  \mathbf{MG}(\mathbf{G}_{l}, \mathbf{F}^{\mathbf{q}}_{l-1}),
\end{equation}
\begin{equation}
\label{pred_func}
        \mathbf{MG}(\mathbf{G},\mathbf{F}) =  Sigmoid(f_G(\mathbf{G}) \mathbf{F}^{\top}),
\end{equation}
where $f_G(\cdot)$ is another linear transformation. 
At the same time, the generated prototype $\mathbf{G}_{l}$ is also used to update the query feature maps via query activation ($\mathbf{QA}$):
\begin{equation}
\label{query_act}
    \mathbf{F}^{\mathbf{q}}_{l} = \mathbf{QA}(\mathbf{G}_{l},\mathbf{F}^{\mathbf{q}}_{l-1}) = \mathcal{F}_{actv} (\mathbf{G}_{l} \circledcirc \mathbf{F}^{\mathbf{q}}_{l-1}),
\end{equation}
where $\circledcirc$ means concatenation and $\mathcal{F}_{actv}$ is an activation network with two convolution layers. 

In the original implementation, IPMT first feeds the input images into a froze backbone encoder (\eg ResNet-50 \cite{he2016deep}), obtaining multi-scale features $\left\{\mathbf{X}_1,\mathbf{X}_2,\mathbf{X}_3,\mathbf{X}_4 \right\}$ from the last four convolution blocks, with the scale of 1/4, 1/8, 1/8, and 1/8, respectively.
Then, $\mathbf{X}_1$ and $\mathbf{X}_2$ are fused to obtain the backbone features for both support and query. Next, a prototype activation (PA) module
generates the support features $\mathbf{F}^{\mathbf{s}}$, the initial query feature $\mathbf{F}^{\mathbf{q}}_0$, and the initial query mask $\tilde{\mathbf{O}}^{\mathbf{q}}_{0}$. The initial prototype $\mathbf{G}_{0}$ is randomly initialized at the
beginning and then optimized during training.
Afterward, by iteratively operating \eqref{ipm} to \eqref{query_act} in five layers, both prototype and features can be optimized step by step.
For more details please refer to \cite{liu2022ipmt}.

\section{Video IPMT}
As shown in Figure~\ref{fig:overview}, we take the original IPMT model as the baseline and combine it with our proposed multi-grained prototype learning scheme for performing the FSVOS task.
we elaborate on the design of our clip-level, frame-level, and memory-level prototypes first. Finally, we present the training loss, including the newly proposed CCDS loss.

\subsection{Clip Prototype Learning}
Since a video is composed of several consecutive frames, a straightforward way is to directly apply the IPMT model on each query frame. However, such a naive method does not consider any video temporal information and also lacks efficiency. 
Hence, we propose to take video clips as query units for both training and inference. Given a query video clip $\mathcal C=\{\mathbf{I}^{\mathbf{c}_i}\}_{i=1}^{T_c}$ with $T_c$ frames, we perform the masked attention with clip-level query features to generate the clip prototype:
\begin{equation}
\label{g_clip}
    \mathbf{G}^{\mathbf{c}}_{l} =  \mathbf{MAtt}(\mathbf{G}_{l-1},\mathbf{F}^{\mathbf{c}}_{l-1},\tilde{\mathbf{O}}^{\mathbf{f}}_{l-1}),
\end{equation}
where $\mathbf{F}^{\mathbf{c}}_{l-1} \in \mathbb{R}^{T_c hw\times C}$, $\tilde{\mathbf{O}}^{\mathbf{f}}_{l-1} \in \mathbb{R}^{T_c hw\times 1}$ is the previous predicted masks and will be explained later. Then, we obtain the clip-level intermediate prototype:
\begin{equation}
\label{ipm_clip}
\begin{aligned}
        \mathbf{G}_{l}^{\mathbf{ci}} =  \mathbf{MLP}(\mathbf{G}^{\mathbf{s}}_{l} + \mathbf{G}^{\mathbf{c}}_{l}+\mathbf{G}_{l-1}),
\end{aligned}
\end{equation}
which encodes local guidance information within the whole query clip. Next, we can use \eqref{pred_func} to segment the target objects in the whole clip and obtain clip predictions $\tilde{\mathbf{O}}^{\mathbf{c}}_{l}\in \mathbb{R}^{T_c hw\times 1}=\mathbf{MG}(\mathbf{G}_{l}^{\mathbf{ci}},\mathbf{F}^{\mathbf{c}}_{l-1})$, which preserves temporal coherence within the $T_c$ frames.

\subsection{Frame Prototype Learning}\label{frame_proto}
The clip-level intermediate prototype $\mathbf{G}_{l}^{\mathbf{ci}}$ uses one prototype to represent the whole clip and encodes the clip consensus object information, however, may ignore frame-level fine-grained cues. This problem may cause a performance drop when the object appearance changes significantly within the clip. To mitigate this problem, we propose to further generate frame-level query prototypes $\{\mathbf{G}_{l}^{\mathbf{f}_i}\}_{i=1}^{T_c}$ by updating $\mathbf{G}_{l}^{\mathbf{ci}}$ with related fine-grained object cues in each frame:
\begin{equation}
\label{g_frame}
    \mathbf{G}_{l}^{\mathbf{f}_i} =  \mathbf{MLP}(\mathbf{MAtt}(\mathbf{G}_{l}^{\mathbf{ci}},\mathbf{F}^{\mathbf{c}_i}_{l-1},\tilde{\mathbf{O}}^{\mathbf{c}_i}_{l})+\mathbf{G}_{l}^{\mathbf{ci}}).
\end{equation}
In this process, we use $\mathbf{G}_{l}^{\mathbf{ci}}$ as a good initialization to learn each $\mathbf{G}_{l}^{\mathbf{f}_i}$ 
and subsequently aggregate the frame-level query prototypes as the initialization for the next iteration:
\begin{equation}
\label{g_frame2clip}
    \mathbf{G}_{l} =  \frac{1}{T_c}\sum_{i = 1}^{T_c}\mathbf{G}_{l}^{\mathbf{f}_i}.
\end{equation}
As such, we enable \emph{bidirectional} clip-frame prototype communication, which promotes the intra-clip temporal correlation. On the contrary, directly using the clip-level intermediate prototype $\mathbf{G}_{l}^{\mathbf{ci}}$ as the initialization for the next iteration only enables \emph{one-way} communication from the clip information to the frame-level.

At the same time, we use each frame prototype $\mathbf{G}_{l}^{\mathbf{f}_i}$ to generate the frame-level segmentation mask and update the query feature for each frame independently:
\begin{equation}
\begin{split}
\label{frame_pred_qa}
        \tilde{\mathbf{O}}^{\mathbf{f}_i}_{l} = \mathbf{MG}(\mathbf{G}_{l}^{\mathbf{f}_i}, \mathbf{F}^{\mathbf{c}_i}_{l-1}),\\
        \mathbf{F}^{\mathbf{c}_i}_{l} = \mathbf{QA}(\mathbf{G}_{l}^{\mathbf{f}_i},\mathbf{F}^{\mathbf{c}_i}_{l-1}).
\end{split}
\end{equation}
Next, $\tilde{\mathbf{O}}^{\mathbf{f}}_{l}=\{\tilde{\mathbf{O}}^{\mathbf{f}_i}_{l}\}_{i=1}^{T_c}$, $\mathbf{F}^{\mathbf{c}}_{l}=\{\mathbf{F}^{\mathbf{c}_i}_{l}\}_{i=1}^{T_c}$, and $\mathbf{G}_{l}$ are input to \eqref{g_clip} again for the next iteration.

\begin{figure}[t]
\centering
    \includegraphics[width=1\linewidth]{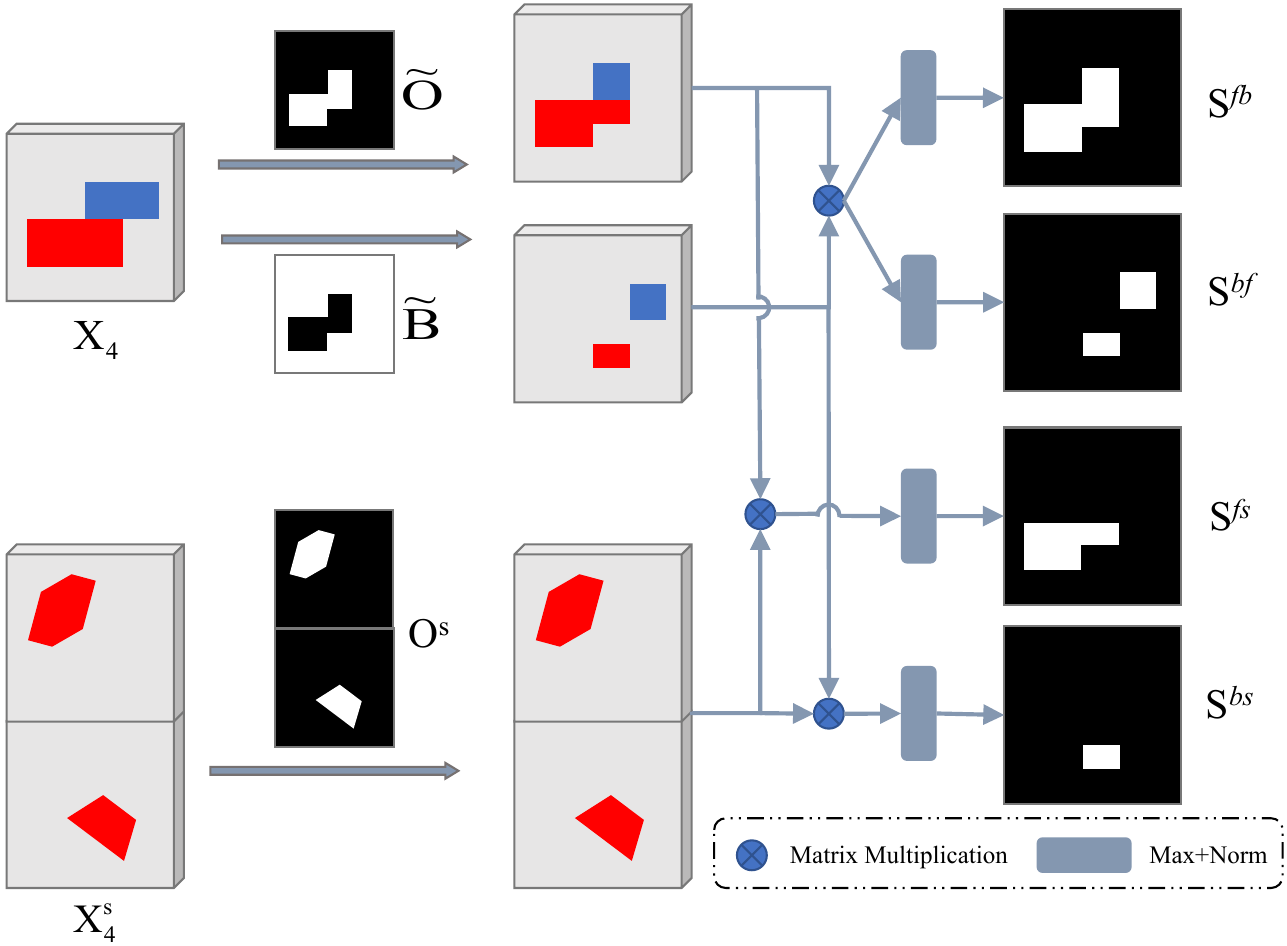}
	\caption{\textbf{Illustration of the generation process of the four structural similarity maps.} Red and blue regions indicate objects of two categories, respectively. Here we only show two support frames for concision.}
	\label{fig:ssm}
	\vspace{-2mm}
\end{figure}

\subsection{Memory Prototype Learning} \label{memory prototype learning}
Since a whole video usually contains numerous frames, only considering the information within a local clip is suboptimal as long-term historical temporal information is ignored. Many VOS works have also demonstrated that leveraging history memory is crucial for video data to enhance the temporal correlation. To this end, we propose to learn a memory prototype for providing historical guidance information. For each clip except the first one, we can use $T_m$ previous frames as the memory set $\mathcal M=\{\mathbf{I}^{\mathbf{m}_i}\}_{i=1}^{T_m}$, which also has predicted masks $\tilde{\mathbf{O}}^{\mathbf{m}} \in \mathbb{R}^{T_m hw\times 1}$. Then, we use masked attention on memory features to generate the memory prototype:
\begin{equation}
\label{g_mem}
    \mathbf{G}^{\mathbf{m}}_{l} =  \mathbf{MAtt}(\mathbf{G}_{l-1},\mathbf{F}^{\mathbf{m}},\tilde{\mathbf{O}}^{\mathbf{m}}),
\end{equation}
where $\mathbf{F}^{\mathbf{m}} \in \mathbb{R}^{T_m hw\times C}$ and $\tilde{\mathbf{O}}^{\mathbf{m}} \in \mathbb{R}^{T_m hw\times 1}$. Then, the update of the clip-level intermediate prototype $\mathbf{G}_{l}^{\mathbf{ci}}$ in \eqref{ipm_clip} can be rewrote as
\begin{equation}
\label{ipm_clip_mem}
\begin{aligned}
        \mathbf{G}_{l}^{\mathbf{ci}} =  \mathbf{MLP}(\mathbf{G}^{\mathbf{s}}_{l} + \mathbf{G}^{\mathbf{c}}_{l}+\mathbf{G}^{\mathbf{m}}_{l}+\mathbf{G}_{l-1}).
\end{aligned}
\end{equation}
We do not consider using memory prototype at the frame level since $\mathbf{G}_{l}^{\mathbf{ci}}$ is designed to combine comprehensive guidance information at the clip-level, simultaneously from support, clip, and memory. In contrast, $\mathbf{G}_{l}^{\mathbf{f}_i}$ encodes pure frame adaptive cues and uses $\mathbf{G}_{l}^{\mathbf{ci}}$ as the initialization.

\vspace{1mm}
\noindent\textbf{Reliable Memory Selection.} During the training stage, we can use the ground truth memory masks $\mathbf{O}^{\mathbf{m}}$ to replace the predicted masks for training an accurate model. However, during the inference stage, we can only use the predicted masks $\tilde{\mathbf{O}}^{\mathbf{m}}$, which may be very noisy and hence heavily contaminate the learned prototype.
To mitigate this problem, we follow \cite{liu2022learning} and train an IoU regression network (IoUNet) to select reliable memory frames which have higher segmentation quality.
Different from \cite{liu2022learning}, 
we explicitly consider the nature of the FSVOS task and propose to leverage the structural similarity among the predicted foreground region, background region, and object region in support images for predicting more accurate IoU scores.

Specifically, as shown in Figure~\ref{fig:ssm}, given the backbone features of an memory frame $\mathbf{X}_4 \in \mathbb{R}^{hw\times C}$ and the support set $\mathbf{X}_4^{\mathbf{s}} \in \mathbb{R}^{Khw\times C}$, the predicted memory mask $\tilde{\mathbf{O}} \in \mathbb{R}^{hw\times 1}$ (here we omit the superscript $\mathbf{m}$ for conciseness), and the ground truth support masks $\mathbf{O}^{\mathbf{s}} \in \mathbb{R}^{Khw\times 1}$, we first obtain the background mask for the memory $\tilde{\mathbf{B}}=1-\tilde{\mathbf{O}}$.
Then, we can evaluate the segmentation quality of the memory frame using a simple and intuitive idea: once it is well segmented, its foreground region $\tilde{\mathbf{O}}$ should be dissimilar to the background $\tilde{\mathbf{B}}$ and similar to the support foreground $\mathbf{O}^{\mathbf{s}}$, while the background $\tilde{\mathbf{B}}$ should be dissimilar to both $\tilde{\mathbf{O}}$ and $\mathbf{O}^{\mathbf{s}}$.
To this end, we compute the structural similarity map of the foreground area of the memory w.r.t. the background area:
\begin{equation}
\label{fb_corr}
    \mathbf{S}^{fb} = \mathbf{Norm}(\mathbf{Max}((\mathbf{X}_4 \odot \tilde{\mathbf{O}})(\mathbf{X}_4 \odot \tilde{\mathbf{B}})^{\top},1)),
\end{equation}
where $\odot$ means element-wise multiplication, $\mathbf{Max}(*,1)$ means obtaining the maximum value along each row, and $\mathbf{Norm}$ denotes the min-max normalization to scale the data to the range of $[0,1]$. Then, the output map $\mathbf{S}^{fb}\in \mathbb{R}^{hw\times 1}$ exactly represents how similar each foreground pixel is w.r.t. the background area, as shown in Figure~\ref{fig:ssm}. 

Likewise, we can obtain a similarity map $\mathbf{S}^{bf}$ of the background area w.r.t. the foreground, similarity map $\mathbf{S}^{fs}$ of the foreground area of the memory w.r.t. the foreground of the support, and similarity map $\mathbf{S}^{bs}$ of the background area of the memory w.r.t. the foreground of the support:
\begin{equation}
\label{other_corr}
\begin{aligned}
        &\mathbf{S}^{bf} = \mathbf{Norm}(\mathbf{Max}((\mathbf{X}_4 \odot \tilde{\mathbf{B}})(\mathbf{X}_4 \odot \tilde{\mathbf{O}})^{\top},1)),\\
        &\mathbf{S}^{fs} = \mathbf{Norm}(\mathbf{Max}((\mathbf{X}_4 \odot \tilde{\mathbf{O}})(\mathbf{X}_4^{\mathbf{s}} \odot \mathbf{O}^{\mathbf{s}})^{\top},1)),\\
        &\mathbf{S}^{bs} = \mathbf{Norm}(\mathbf{Max}((\mathbf{X}_4 \odot \tilde{\mathbf{B}})(\mathbf{X}_4^{\mathbf{s}} \odot \mathbf{O}^{\mathbf{s}})^{\top},1)).\\
\end{aligned}
\end{equation}

Once obtain the four structural similarity maps, we combine them with multi-scale memory features and the original predicted mask $\tilde{\mathbf{O}}$ to regress the IoU score of the memory frame. Concretely, we first downsample $\mathbf{X}_1$ to the 1/8 scale with stride convolution and then fuse it with $\mathbf{X}_2,\mathbf{X}_3,\mathbf{X}_4$ into a combined feature with 256 channels. Then, we concatenate the feature with the four structural similarity maps and $\tilde{\mathbf{O}}$ and input them into four convolution layers and two fully connected layers for predicting the IoU score.

During training, we design three ways to train the IoUNet. First, we can jointly train it with the FSVOS model, in which case real predicted masks can be used for training the IoUNet. Second, we can train it independently using synthesized video data. Specifically, we randomly add noises to the ground truth masks in the training set and use the synthesized masks to mimic good and bad predictions. 
In the third way, we train the IoUNet using synthesized image data with the same noisy mask strategy.
We compute the ground truth IoU score for each training mask and use a $L_1$ loss as supervision. 
During testing, we simply select reliable memory frames whose predicted IoU scores are larger than a threshold.

\subsection{Training Loss}
We follow IPMT \cite{liu2022ipmt} and use the final query feature $\mathbf{F}^{\mathbf{c}}_{5}$ to predict the final segmentation masks via two convolution layers, where Dice loss and IoU loss are both used for optimization. To ease the network training, in each iteration we use binary cross entropy (BCE) loss on the predicted support masks $\tilde{\mathbf{O}}^{\mathbf{s}}_{l}$, clip masks $\tilde{\mathbf{O}}^{\mathbf{c}}_{l}$, and frame masks $\tilde{\mathbf{O}}^{\mathbf{f}}_{l}$.
As for memory frames, 
we generate predictions for them at each iteration using $\mathbf{MG}(\mathbf{G}_{l}^{\mathbf{ci}},\mathbf{F}^{\mathbf{m}})$, and then adopt the BCE loss computed with their ground truth masks as supervision.

\vspace{1mm}
\noindent\textbf{Cross Category Discriminative Segmentation Loss.} To make the learned clip-level intermediate prototypes $\mathbf{G}^{\mathbf{ci}}$ more category discriminative, we furthermore propose a CCDS loss and adopt it within each batch. Suppose our batchsize is set to $B>1$, we select $B$ video clips with different object categories to form each batch. Then, during training, we use the prototypes of each video to perform segmentation on other videos and let the predicted masks be all zeros. The idea behind is intuitive: if the prototypes are well optimized for a specific category, they should not activate any region on other videos with different categories. However, since a video may contain multiple object categories and we only have the ground truth of one class under the one-way few-shot learning setting, we propose to use the ground truth masks to filter out the regions for loss calculation:
\begin{eqnarray}
\label{ccds_loss}
    L_{ccds} & = & \frac{1}{(B-1)B}\sum_{b=1}^{B}\sum_{j\neq b}\frac{1}{\sum \mathbf{O}_{j}} L_{b,j},\\
    L_{b,j} & = & \sum_{l=1}^{5}BCE(\mathbf{MG}(\mathbf{G}_{l,b}^{\mathbf{ci}},\mathbf{F}_{j})\odot \mathbf{O}_{j}, \mathbf{Z}),
\end{eqnarray}
where $\mathbf{G}^{\mathbf{ci}}_{l,b}$ is the clip-level intermediate prototype in the $l$-th iteration of the $b$-th video, $\mathbf{O}_{j}$ and $\mathbf{F}_{j}$ are the ground truth masks and the backbone features of the $j$-th video. $\mathbf{Z}$ is an all zero matrix.

\section{Experiments}
\subsection{Datasets and Evaluation Setting} \label{section4.1}
\noindent\textbf{Datasets.}
Following \cite{chen2021danet,siam2022tti}, our experiments are conducted on the training set of YouTube-VIS \cite{yang2019video} dataset, which has 40 categories and contains 2238 videos with 3774 instances. The dataset is divided into four folds, following the same slit with \cite{chen2021danet,siam2022tti}. In each fold, we use 30 categories for training and the rest 10 categories for testing. Following \cite{chen2021danet,siam2022tti}, we set our experiments on the 5-shot setting and randomly select five images from different videos of the same category as the support set.
We also follow \cite{siam2022tti} and adopt the MiniVSPW \cite{siam2022tti} dataset to evaluate the generalizability of FSVOS methods. It includes 20 categories and provides longer video sequences than YouTube-VIS, since being more challenging.

\vspace{1mm}
\noindent\textbf{Evaluation Setting.} 
As the prior work \cite{chen2021danet} and VOS methods, we adopt the region similarity $\mathcal{J}$ and contour similarity $\mathcal{F}$ for performance evaluation. 
Apart from this, we also follow \cite{siam2022tti} and consider the video consistency metric $\text{VC}_{7}$, which captures temporal prediction consistency among long-range adjacent frames over a temporal window of 7. We adopt the average score on four folds, \ie $\text{mVC}_{7}$, to evaluate the overall performance. We also follow \cite{chen2021danet,siam2022tti} and run the evaluation process five times and report the average results.

Furthermore, there are two different evaluation protocols proposed in  \cite{chen2021danet} and \cite{siam2022tti} for sampling episodes during testing. \emph{Protocol I} fixes the sampled support set for all query videos belonging to the same class in each run, while \emph{protocol II} randomly samples a support set for every query video, which ensures a more stable performance evaluation. Hence, \emph{in this paper we adopt protocol II for evaluation.}

\vspace{1mm}
\noindent\textbf{Implementation Details.}
We adopt ResNet-50 \cite{he2016deep} pretrained on ImageNet \cite{russakovsky2015imagenet} as our encoder backbone. Following IPMT \cite{liu2022ipmt}, we freeze the parameters of the backbone during training and do not adopt online finetuning as \cite{chen2021danet,siam2022tti} did. The iteration step is set to 5 as IPMT.
We set the clip length $T_c$ and the memory length $T_m$ to 5 without further tuning. The IoU threshold for selecting reliable memory frames is set to 0.8. When the reliable memory has more frames, we simply randomly select five frames from them. During training, we randomly select three clips from a video as a sample, where the first clip is trained without using memory, the second clip uses the first clip as memory, and the third clip randomly selects five frames from the first two as the memory. We also share the $\mathbf{MAtt}$ operation for support and memory since we found this leads to better performance.

We use Adam as our optimizer. The batchsize is set to $B=4$ and the learning rate is set to 5e-4. We train our model 100 epochs in total. All experiments are conducted on a NVIDIA Tesla A100 GPU. We adopt random horizontal flip, random crop and random resize to augment the training data. During training and testing, all video frames are downsampled to the resolution of (240,424) as the inputs.

\begin{table*}[t]
	\centering
	\resizebox{\linewidth}{!}{
\begin{tabular}{@{}ccccccccccccc@{}}
\toprule

\multicolumn{1}{c|}{}       & \multicolumn{1}{c|}{}            & \multicolumn{5}{c|}{$\mathcal{J}$}                & \multicolumn{5}{c|}{$\mathcal{F}$}    & \multicolumn{1}{c}{}                 \\

 \multicolumn{1}{c|}{\multirow{-2}{*}{Methods}} & \multicolumn{1}{c|}{\multirow{-2}{*}{Name}}          & Fold-1                       & Fold-2                       & Fold-3                       & Fold-4                       & \multicolumn{1}{c|}{ \textbf{Mean}}                         & Fold-1                    & Fold-2                    & Fold-3                    & Fold-4                    &  \multicolumn{1}{c|}{\textbf{Mean}}                   &    \multicolumn{1}{c}{\multirow{-2}{*}{$\text{mVC}_{7}$}}               \\ \midrule
 
    \multicolumn{1}{c|}{}   &\multicolumn{1}{c|}{NTRENet \cite{liu2022ntrenet}}   & 
    \multicolumn{1}{c}{39.0}  & \multicolumn{1}{c}{66.4}  &  \multicolumn{1}{c}{61.7} &  \multicolumn{1}{c}{61.2}  &  \multicolumn{1}{c}{57.1}  & \multicolumn{1}{c}{41.1}  & \multicolumn{1}{c}{63.0}  &  \multicolumn{1}{c}{60.1}  &  \multicolumn{1}{c}{59.4}  &  \multicolumn{1}{c}{55.9} &  \multicolumn{1}{c}{55.7} \\ 
        
    \multicolumn{1}{c|}{}   &\multicolumn{1}{c|}{SSP \cite{fan2022ssp}}   & 
    \multicolumn{1}{c}{46.7}  & \multicolumn{1}{c}{64.3}  &  \multicolumn{1}{c}{59.3} &  \multicolumn{1}{c}{54.5}  &  \multicolumn{1}{c}{56.2}  & \multicolumn{1}{c}{34.8}  & \multicolumn{1}{c}{53.0}  &  \multicolumn{1}{c}{46.0} &  \multicolumn{1}{c}{46.7}  &  \multicolumn{1}{c}{45.1}  &  \multicolumn{1}{c}{49.6} \\

    \multicolumn{1}{c|}{}   &\multicolumn{1}{c|}{VAT \cite{hong2022cost}}   & 
    \multicolumn{1}{c}{42.6}  & \multicolumn{1}{c}{62.8}  &  \multicolumn{1}{c}{57.0} &  \multicolumn{1}{c}{56.7}  &  \multicolumn{1}{c}{54.8}  & \multicolumn{1}{c}{41.6}  & \multicolumn{1}{c}{56.4}  &  \multicolumn{1}{c}{50.8} &  \multicolumn{1}{c}{53.0}  &  \multicolumn{1}{c}{50.5}  &  \multicolumn{1}{c}{54.9} \\

    \multicolumn{1}{c|}{\multirow{-4}{*}{FSISS}}   &\multicolumn{1}{c|}{IPMT \cite{liu2022ipmt}}   & 
    \multicolumn{1}{c}{43.8}  & \multicolumn{1}{c}{65.8}  &  \multicolumn{1}{c}{61.0} &  \multicolumn{1}{c}{60.7}  &  \multicolumn{1}{c}{57.8}  & \multicolumn{1}{c}{42.5}  & \multicolumn{1}{c}{59.5}  &  \multicolumn{1}{c}{57.8} &  \multicolumn{1}{c}{55.9}  &  \multicolumn{1}{c}{53.9}  &  \multicolumn{1}{c}{57.7} \\ \midrule

    \multicolumn{1}{c|}{}   & \multicolumn{1}{c|}{DAN \cite{chen2021danet}}   & \multicolumn{1}{c}{43.9}  & \multicolumn{1}{c}{64.5}  &  \multicolumn{1}{c}{61.1} &  \multicolumn{1}{c}{62.1}  &  \multicolumn{1}{c}{57.9}  & \multicolumn{1}{c}{42.4}  & \multicolumn{1}{c}{62.0}  &  \multicolumn{1}{c}{60.0} &  \multicolumn{1}{c}{60.0}  &  \multicolumn{1}{c}{56.1}  &  \multicolumn{1}{c}{41.5} \\ 
    
    \multicolumn{1}{c|}{}   &\multicolumn{1}{c|}{TTI \cite{siam2022tti}}   & \multicolumn{1}{c}{47.2}  & \multicolumn{1}{c}{68.8}  &  \multicolumn{1}{c}{61.4} &  \multicolumn{1}{c}{63.5}  &  \multicolumn{1}{c}{60.2}  & \multicolumn{1}{c}{-}  & \multicolumn{1}{c}{-}  &  \multicolumn{1}{c}{-} &  \multicolumn{1}{c}{-}  &  \multicolumn{1}{c}{-}  &  \multicolumn{1}{c}{60.8} \\

\multicolumn{1}{c|}{\multirow{-4}{*}{FSVOS}}    &  \multicolumn{1}{c|}{\cellcolor[HTML]{C0C0C0} VIPMT(Ours)}                 & \cellcolor[HTML]{C0C0C0}\textbf{{50.6}} & 
\cellcolor[HTML]{C0C0C0}\textbf{{70.9}} & 
\cellcolor[HTML]{C0C0C0}\textbf{{68.8}}  & 
\cellcolor[HTML]{C0C0C0}\textbf{{66.5}} & 
\cellcolor[HTML]{C0C0C0}\textbf{{64.2}} & 
\cellcolor[HTML]{C0C0C0}\textbf{{51.3}} & 
\cellcolor[HTML]{C0C0C0}\textbf{{66.9}} & 
\cellcolor[HTML]{C0C0C0}\textbf{{65.2}} & 
\cellcolor[HTML]{C0C0C0}\textbf{{64.4}} & 
\cellcolor[HTML]{C0C0C0}\textbf{{62.0}} & 
\cellcolor[HTML]{C0C0C0}\textbf{{65.7}} \\ 
\bottomrule

\end{tabular}}
\caption{\textbf{Comparison with state-of-the-art methods on YouTube-VIS \cite{yang2019video}\protect\footnotemark.} \textbf{Bold} means the best performance.}
\label{ytvis_compar}
\vspace{-2mm}
\end{table*}

\begin{table*}[t]
	\centering
	\resizebox{\linewidth}{!}{
\begin{tabular}{@{}ccccccccccccc@{}}
\toprule

\multicolumn{1}{c|}{}       & \multicolumn{1}{c|}{}            & \multicolumn{5}{c|}{$\mathcal{J}$}                & \multicolumn{5}{c|}{$\mathcal{F}$}    & \multicolumn{1}{c}{}                 \\

 \multicolumn{1}{c|}{\multirow{-2}{*}{Methods}} & \multicolumn{1}{c|}{\multirow{-2}{*}{Name}}          & Fold-1                       & Fold-2                       & Fold-3                       & Fold-4                       & \multicolumn{1}{c|}{ \textbf{Mean}}                         & Fold-1                    & Fold-2                    & Fold-3                    & Fold-4                    &  \multicolumn{1}{c|}{\textbf{Mean}}                   &    \multicolumn{1}{c}{\multirow{-2}{*}{$\text{mVC}_{7}$}}               \\ \midrule

    \multicolumn{1}{c|}{}   &\multicolumn{1}{c|}{TTI \cite{siam2022tti}}   & \multicolumn{1}{c}{25.2}  & \multicolumn{1}{c}{37.1}  &  \multicolumn{1}{c}{25.0} &  \multicolumn{1}{c}{\textbf{29.6}}  &  \multicolumn{1}{c}{29.2}  & \multicolumn{1}{c}{-}  & \multicolumn{1}{c}{-}  &  \multicolumn{1}{c}{-} &  \multicolumn{1}{c}{-}  &  \multicolumn{1}{c}{-}  &  \multicolumn{1}{c}{24.4} \\

\multicolumn{1}{c|}{\multirow{-2}{*}{FSVOS}}    &  \multicolumn{1}{c|}{\cellcolor[HTML]{C0C0C0} VIPMT(Ours)}                 & \cellcolor[HTML]{C0C0C0}\textbf{{26.2}} & 
\cellcolor[HTML]{C0C0C0}\textbf{{42.2}} & 
\cellcolor[HTML]{C0C0C0}\textbf{{31.6}} & \cellcolor[HTML]{C0C0C0}{29.4} & \cellcolor[HTML]{C0C0C0}\textbf{{32.4}}           & \cellcolor[HTML]{C0C0C0}\textbf{{30.6}} & \cellcolor[HTML]{C0C0C0}\textbf{{45.7}}   & \cellcolor[HTML]{C0C0C0}\textbf{{36.3}} & \cellcolor[HTML]{C0C0C0}\textbf{{34.2}} & \cellcolor[HTML]{C0C0C0}\textbf{{36.7}} & \cellcolor[HTML]{C0C0C0}{\textbf{42.1}} \\ \bottomrule

\end{tabular}}
\caption{\textbf{Comparison with TTI \cite{siam2022tti} on MiniVSPW \cite{siam2022tti}.}}
\label{Voccompar_minivspw}
\vspace{-3mm}
\end{table*}

\subsection{Comparison with State-of-the-art Methods} \label{section4.2}
In Table~\ref{ytvis_compar}, we compare our VIPMT with four recently published FSISS methods and two FSVOS methods, on the YouTuebe-VIS \cite{yang2019video} dataset.
We find that our VIPMT largely improves all metrics, \ie 4\% $\mathcal{J}$ mean score, near 6\% $\mathcal{F}$ mean score, and near 5\% $\text{mVC}_{7}$ score, although previous works \cite{chen2021danet,siam2022tti} used on-line learning while we didn't.

\footnotetext{Please note that here we re-tested the performance of DAN \cite{chen2021danet} using \emph{protocol II}. Hence, its scores are different from the original paper. For TTI \cite{siam2022tti}, since the authors did not provide the scores for $\mathcal{F}$, we leave them blank. We retrained all FSISS methods on YouTube-VIS for a fair comparison.}

To verify the generalization ability of our method, we compare our method with TTI \cite{siam2022tti} on MiniVSPW.  Table~\ref{Voccompar_minivspw} shows our method obtains a significant improvement of more than 3\% on $\mathcal{J}$ and more than 17\% on $\text{mVC}_{7}$, which verifies that our method works well in more challenging scenarios.

\begin{figure*}[!t]
    \centering
    \includegraphics[width=1\linewidth]{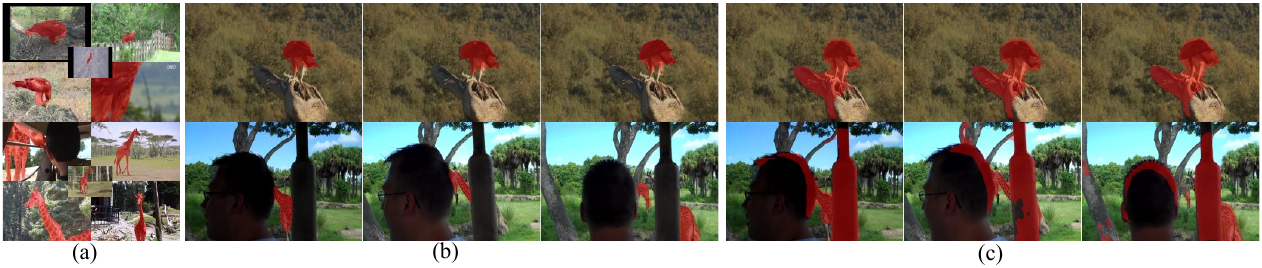}
    \caption{\textbf{Visualization of comparison cases.} (a) Support sets. (b) Predicted masks of our VIMPT. (c) Predicted masks from TTI \cite{siam2022tti}.} 
    \label{sota_visualize}
\end{figure*}

In Figure~\ref{sota_visualize}, we give some visual comparison cases. Compared with TTI \cite{siam2022tti}, our VIMPT is less distracted by other objects and achieves more precise segmentation, even in highly occluded scenes (bottom row).

We also include three state-of-the-art semi-supervised VOS methods for comparison, \ie STCN \cite{cheng2021stcn}, XMem \cite{cheng2022xmem}, and RDE-VOS \cite{li2022rde}. Specifically, we use IPMT for segmenting the first frame and then use VOS methods for propagating the segmentation to the full video. Table~\ref{vos_compar} demonstrates that such an ad-hoc combination of FSISS and VOS methods can not achieve as good results as our model does.

\begin{table}[t]
	\centering
    \small
\scalebox{0.8}{
\begin{tabular}{@{}c|cccc@{}}
\toprule
\multicolumn{1}{c|}{\multirow{-1}{*}{Datasets}} & \multicolumn{1}{c|}{Methods}    & \multicolumn{1}{c|}{$\mathcal{J}$-Mean}  & \multicolumn{1}{c|}{$\mathcal{F}$-Mean}   & \multicolumn{1}{c}{$\text{mVC}_{7}$}        \\ \midrule

\multicolumn{1}{c|}{ } 
& \multicolumn{1}{c|}{IPMT+STCN \cite{cheng2021stcn}}  & \multicolumn{1}{c}{58.9}  & \multicolumn{1}{c}{56.5}   & \multicolumn{1}{c}{61.7} 
\\

\multicolumn{1}{c|}{} 
& \multicolumn{1}{c|}{IPMT+XMem \cite{cheng2022xmem}}  & \multicolumn{1}{c}{59.2}  & \multicolumn{1}{c}{56.8}   & \multicolumn{1}{c}{62.2}   \\

\multicolumn{1}{c|}
{\multirow{-2}{*}{YouTube-VIS}}
& \multicolumn{1}{c|}{IPMT+RDE-VOS \cite{li2022rde}}  
& \multicolumn{1}{c}{59.3} 
& \multicolumn{1}{c}{58.0} 
& \multicolumn{1}{c}{63.3} 
\\

\multicolumn{1}{c|}{} 
& \multicolumn{1}{c|}{VIPMT(Ours)}  
& \multicolumn{1}{c}{\textbf{64.2}} 
& \multicolumn{1}{c}{\textbf{62.0}} 
& \multicolumn{1}{c}{\textbf{65.7}} 
\\ \midrule

\multicolumn{1}{c|}{ } 
& \multicolumn{1}{c|}{IPMT+STCN \cite{cheng2021stcn}}  & \multicolumn{1}{c}{27.7}  & \multicolumn{1}{c}{31.1}   & \multicolumn{1}{c}{39.3} 
\\

\multicolumn{1}{c|}{} 
& \multicolumn{1}{c|}{IPMT+XMem \cite{cheng2022xmem}}  & \multicolumn{1}{c}{27.7}  & \multicolumn{1}{c}{31.7}   & \multicolumn{1}{c}{41.1}   \\

\multicolumn{1}{c|}
{\multirow{-2}{*}{MiniVSPW}}
& \multicolumn{1}{c|}{IPMT+RDE-VOS \cite{li2022rde}}  
& \multicolumn{1}{c}{28.8} 
& \multicolumn{1}{c}{33.5} 
& \multicolumn{1}{c}{40.1} 
\\

\multicolumn{1}{c|}{} 
& \multicolumn{1}{c|}{VIPMT(Ours)}  
& \multicolumn{1}{c}{\textbf{32.4}} 
& \multicolumn{1}{c}{\textbf{36.7}} 
& \multicolumn{1}{c}{\textbf{42.1}} 

\\\bottomrule
\end{tabular}
}
\caption{\textbf{Comparison with state-of-the-art VOS methods.}}
\label{vos_compar}
\vspace{-3mm}
\end{table}

\begin{table}[t]
	\centering
	\resizebox{\linewidth}{!}{
\begin{tabular}{@{}c|cccc|ccc@{}}
\toprule

 \multicolumn{1}{c|}{}     & \multicolumn{4}{c|}{\textbf{Settings}}       & \multicolumn{1}{c|}{}        & \multicolumn{1}{c|}{}    & \multicolumn{1}{c}{}                 \\ 

\multicolumn{1}{c|}{\multirow{-2}{*}{Name}}    & \multicolumn{1}{c}{Clip}  & \multicolumn{1}{c}{Frame}   & \multicolumn{1}{c}{Memory}      & \multicolumn{1}{c|}{CCDS}   & \multicolumn{1}{c|}{\multirow{-2}{*}{$\mathcal{J}$-Mean}}  & \multicolumn{1}{c|}{\multirow{-2}{*}{$\mathcal{F}$-Mean}}   & \multicolumn{1}{c}{\multirow{-2}{*}{$\text{mVC}_{7}$}}              \\ \midrule

\multicolumn{1}{c|}{Baseline}  & \multicolumn{1}{c}{}  & \multicolumn{1}{c}{}   & \multicolumn{1}{c}{}      & \multicolumn{1}{c|}{}   & \multicolumn{1}{c}{57.8}  & \multicolumn{1}{c}{53.9}   & \multicolumn{1}{c}{57.8}          \\ 

\multicolumn{1}{c|}{+C}  & \multicolumn{1}{c}{\ding{52}}  & \multicolumn{1}{c}{}   & \multicolumn{1}{c}{}      & \multicolumn{1}{c|}{}   & \multicolumn{1}{c}{61.8}  & \multicolumn{1}{c}{58.8}   & \multicolumn{1}{c}{60.7}          \\ 

\multicolumn{1}{c|}{+C+F}  &\multicolumn{1}{c}{\ding{52}}  & \multicolumn{1}{c}{\ding{52}}   & \multicolumn{1}{c}{}      & \multicolumn{1}{c|}{}   & \multicolumn{1}{c}{62.5}  & \multicolumn{1}{c}{60.3}   & \multicolumn{1}{c}{62.1}          \\

\multicolumn{1}{c|}{+C+F+M}  &\multicolumn{1}{c}{\ding{52}}  & \multicolumn{1}{c}{\ding{52}}   & \multicolumn{1}{c}{\ding{52}}      & \multicolumn{1}{c|}{}   & \multicolumn{1}{c}{63.5}  & \multicolumn{1}{c}{61.9}   & \multicolumn{1}{c}{62.8}          \\ 

\multicolumn{1}{c|}{VIPMT}  & \multicolumn{1}{c}{\ding{52}}  & \multicolumn{1}{c}{\ding{52}}   & \multicolumn{1}{c}{\ding{52}}      & \multicolumn{1}{c|}{\ding{52}}   & \multicolumn{1}{c}{\textbf{64.2}}  & \multicolumn{1}{c}{\textbf{62.0}}   & \multicolumn{1}{c}{\textbf{65.7}}          \\\bottomrule

\end{tabular}}
\caption{\textbf{Ablation study on the effectiveness of each model component.} Baseline means IPMT \cite{liu2022ipmt}.}
\label{ablationstudy}
\vspace{-3mm}
\end{table}

\subsection{Ablation Study}
In this section, we report ablation study results on the YouTube-VIS dataset. For $\mathcal{J}$ and $\mathcal{F}$ we use their average on four folds, \ie $\mathcal{J}$-Mean and $\mathcal{F}$-Mean.

\vspace{1mm}
\noindent\textbf{Effectiveness of Each Component.}
In Table~\ref{ablationstudy}, we first adopt IPMT \cite{liu2022ipmt} as the baseline, which processes each video frame independently.
Then, we adopt our clip prototype learning, denoted as ``+C". Results show that ``+C" surpasses the baseline by a large margin on all metrics, which verifies the effectiveness of using local temporal guidance with our clip prototype learning. After that, we further add the frame prototype learning to ``+C", resulting in ``+C+F". Benefiting from the frame-level fine-grained cues, ``+C+F" achieves large performance improvement for $\mathcal{F}$-Mean and $\text{mVC}_{7}$. Moreover, we adopt ``+C+F+M" to represent adding the memory prototype learning on ``+C+F". It largely improves the performance for both $\mathcal{J}$-Mean and $\mathcal{F}$-Mean, which shows the importance of long-term temporal information brought by the memory prototype learning. Based on this, we further add the CCDS loss to obtain our final model VIPMT. It also brings obvious improvements, especially on $\text{mVC}_{7}$, from 62.8 to 65.7. This proves the effectiveness of increasing category discrimination for FSVOS.

\begin{figure*}[!t]
    \centering
    \includegraphics[width=1\linewidth]{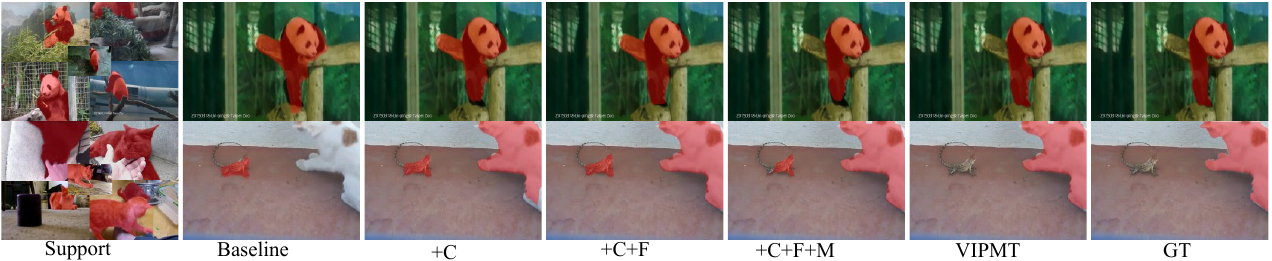}
    \caption{\textbf{Visualization of the comparison of different ablation study models.} We can see progressive improvements brought by the multi-grained temporal prototypes and the CCDS loss.} 
    \label{ablation_visualize}
    \vspace{-2mm}
\end{figure*}

In Figure~\ref{ablation_visualize}, we visualize the predicted masks from the different models mentioned above. We can find that progressive improvements can be brought by using
multi-grained temporal prototypes and the CCDS loss.

\vspace{1mm}
\noindent\textbf{Clip-frame Communication.}
As mentioned in Section~\ref{frame_proto}, we use the mean of the frame prototypes (\eqref{g_frame2clip}) to initialize the intermediate prototype for the next iteration, which enables \emph{bidirectional} clip-frame communication. On the contrary, directly using the clip-level intermediate prototype $\mathbf{G}_{l}^{\mathbf{ci}}$ as the initialization only enables \emph{one-way} communication. Here we compare the effectiveness of the two schemes.
Results in Table~\ref{Initialization of Prototypes} demonstrate that our design of using bidirectional clip-frame communication surpasses the one-way scheme by a large margin on all metrics.

\begin{table}[t]
	\centering
    \small
\begin{tabular}{@{}c|ccc@{}}
\toprule
\multicolumn{1}{c|}{Clip-frame Communication}    & \multicolumn{1}{c|}{$\mathcal{J}$-Mean}  & \multicolumn{1}{c|}{$\mathcal{F}$-Mean}   & \multicolumn{1}{c}{$\text{mVC}_{7}$}              \\ \midrule
\multicolumn{1}{c|}{One-way}   & \multicolumn{1}{c}{61.4}  & \multicolumn{1}{c}{59.1}   & \multicolumn{1}{c}{60.1}          \\
\multicolumn{1}{c|}{Bidirectional}   & \multicolumn{1}{c}{\textbf{62.5}}  & \multicolumn{1}{c}{\textbf{60.3}}   & \multicolumn{1}{c}{\textbf{62.1}}          \\\bottomrule
\end{tabular}
\caption{\textbf{Comparison of two clip-frame communication schemes.}}
\label{Initialization of Prototypes}
\vspace{-4mm}
\end{table}

\vspace{1mm}
\noindent\textbf{Different Training Strategies of IoUNet.}
As mentioned in Section~\ref{memory prototype learning}, we can use three ways to train the IoUNet, \ie using real predicted masks (``Real"), synthesized video data (``Video"), and synthesized image data (``Image"), respectively. For ``Image", we use an FSISS dataset COCO-20$^i$ \cite{nguyen2019feature} and remove the classes overlapped with our video data for IoUNet training. We compare their performance in Table~\ref{strategies_IoUNet}.
We also compare the performance of using our proposed structural similarity maps or not (``w/ SSM" and ``w/o SSM").
The results are reported under the IoU threshold of 0.5.
The table demonstrates that using synthesized image data results in the best performance in both "w/ SSM" and "w/o SSM" settings and using the structural similarity maps always leads to better performance. Another benefit of using image data is that we can train a unified IoUNet for all folds instead of training an IoUNet for each fold. Hence, finally, we use synthesized image data for IoUNet training.

\vspace{1mm}
\noindent\textbf{IoU Threshold.}
Finally, we investigate the best IoU threshold for memory selection. Table~\ref{threshold} indicates that 0.8 is the best threshold and using the proposed structural similarity maps always leads to better performance.

\begin{table}[]
	\centering
        \small
	\resizebox{\linewidth}{!}{
\begin{tabular}{@{}c|ccc|ccc@{}}
\toprule
 \multicolumn{1}{c|}{}     & \multicolumn{3}{c|}{\textbf{w/o SSM}}       & \multicolumn{3}{c}{\textbf{w/ SSM}}                      \\ 

\multicolumn{1}{c|}{\multirow{-2}{*}{Data}}    & \multicolumn{1}{c}{$\mathcal{J}$-Mean}  & \multicolumn{1}{c}{$\mathcal{F}$-Mean}   & \multicolumn{1}{c|}{$\text{mVC}_{7}$}      & \multicolumn{1}{c}{$\mathcal{J}$-Mean}  & \multicolumn{1}{c}{$\mathcal{F}$-Mean}   & \multicolumn{1}{c}{$\text{mVC}_{7}$}           \\ \midrule

\multicolumn{1}{c|}{Real}  & \multicolumn{1}{c}{62.8} & \multicolumn{1}{c}{61.5}   & \multicolumn{1}{c|}{62.6}   & \multicolumn{1}{c}{63.2} & \multicolumn{1}{c}{61.6}   & \multicolumn{1}{c}{62.8}         \\

\multicolumn{1}{c|}{Video}  & \multicolumn{1}{c}{62.9} & \multicolumn{1}{c}{61.6}   & \multicolumn{1}{c|}{62.6}  & \multicolumn{1}{c}{63.2} & \multicolumn{1}{c}{61.6}   & \multicolumn{1}{c}{62.8}          \\

\multicolumn{1}{c|}{Image}  & \multicolumn{1}{c}{63.0} & \multicolumn{1}{c}{\textbf{61.7}}   & \multicolumn{1}{c|}{62.7}    & \multicolumn{1}{c}{\textbf{63.3}} & \multicolumn{1}{c}{\textbf{61.7}}   & \multicolumn{1}{c}{\textbf{63.0}}        \\ \midrule
\end{tabular}}
\caption{\textbf{Different training strategies of IoUNet.} SSM means the proposed structural similarity maps.}
\label{strategies_IoUNet}
\end{table}

\begin{table}[]
	\centering
        \small
	\resizebox{\linewidth}{!}{
\begin{tabular}{@{}c|ccc|ccc@{}}
\toprule

 \multicolumn{1}{c|}{}     & \multicolumn{3}{c|}{\textbf{w/o SSM}}       & \multicolumn{3}{c}{\textbf{w/ SSM}}                      \\ 

\multicolumn{1}{c|}{\multirow{-2}{*}{Threshold}}    & \multicolumn{1}{c}{$\mathcal{J}$-Mean}  & \multicolumn{1}{c}{$\mathcal{F}$-Mean}   & \multicolumn{1}{c|}{$\text{mVC}_{7}$}      & \multicolumn{1}{c}{$\mathcal{J}$-Mean}  & \multicolumn{1}{c}{$\mathcal{F}$-Mean}   & \multicolumn{1}{c}{$\text{mVC}_{7}$}           \\ \midrule

\multicolumn{1}{c|}{0.5}  & \multicolumn{1}{c}{63.0} & \multicolumn{1}{c}{61.7}   & \multicolumn{1}{c|}{62.7}   & \multicolumn{1}{c}{63.3} & \multicolumn{1}{c}{61.7}   & \multicolumn{1}{c}{\textbf{63.0}}         \\

\multicolumn{1}{c|}{0.6}  & \multicolumn{1}{c}{63.1} & \multicolumn{1}{c}{61.7}   & \multicolumn{1}{c|}{62.9}  & \multicolumn{1}{c}{63.3} & \multicolumn{1}{c}{61.7}   & \multicolumn{1}{c}{62.7}          \\

\multicolumn{1}{c|}{0.7}  & \multicolumn{1}{c}{63.1} & \multicolumn{1}{c}{61.8}   & \multicolumn{1}{c|}{62.9}    & \multicolumn{1}{c}{63.4} & \multicolumn{1}{c}{61.8}   & \multicolumn{1}{c}{62.6}        \\

\multicolumn{1}{c|}{0.8}  & \multicolumn{1}{c}{63.1} & \multicolumn{1}{c}{61.8}   & \multicolumn{1}{c|}{62.5}    & \multicolumn{1}{c}{\textbf{63.5}} & \multicolumn{1}{c}{\textbf{61.9}}   & \multicolumn{1}{c}{62.8}        \\ 

\multicolumn{1}{c|}{0.9}  & \multicolumn{1}{c}{63.0} & \multicolumn{1}{c}{61.8}   & \multicolumn{1}{c|}{62.3}    & \multicolumn{1}{c}{\textbf{63.5}} & \multicolumn{1}{c}{\textbf{61.9}}   & \multicolumn{1}{c}{62.5}        \\ \bottomrule

\end{tabular}}
\caption{\textbf{Experimental results of using different IoU thresholds. }}
\label{threshold}
\vspace{-3mm}
\end{table}

\section{Conclusion}
We explored learning multi-grained temporal prototypes to tackle the FSVOS task. 
Based on the IPMT model, the query video information was decomposed into a clip prototype, a memory prototype, and frame prototypes.
We also improved an IoU regression method for selecting reliable memory for FSVOS by leveraging the task prior knowledge
and proposed a new loss to enhance the category discriminability of the prototypes. The effectiveness of our proposed VIPMT has been verified on two benchmark datasets.

\vspace{-4mm}
\small{
\paragraph{Acknowledgments:}
This work was supported in part by the National Natural Science Foundation of China under Grants 62071388, 62136007,U20B2065 and 62036005, and the Fundamental Research Funds for the Central Universities under Grant D5000230057.
}

{\small
\bibliographystyle{ieee_fullname}
\bibliography{egbib}
}

\end{document}